# Deep Q-Network Based Decision Making for Autonomous Driving

Max Peter Ronecker[†]
Technical University of Munich
Department of Electrical and Computer Engineering
Munich, Germany
max.ronecker@tum.de

Yuan Zhu
Tongji University
Sino-German College for Graduate Studies[†]
Shanghai, China
yuan.zhu@tongji.edu.cn

*Abstract*—Currently decision making is one of the biggest challenges in autonomous driving. This paper introduces a method for safely navigating an autonomous vehicle in highway scenarios by combining deep Q-Networks and insight from control theory. A Deep Q-Network is trained in simulation to serve as a central decision-making unit by proposing targets for a trajectory planner. The generated trajectories in combination with a controller for longitudinal movement are used to execute lane change maneuvers. In order to prove the functionality of this approach it is evaluated on two different highway traffic scenarios. Furthermore, the impact of different state representations on the performance and training process is analyzed. The results show that the proposed system can produce efficient and safe driving behavior.

## I. Introduction

Autonomous driving is one of the most researched field of our time in both industry and academia. There has been rapid improvement across all fields so that self-driving vehicles do not seem to be far future anymore. First robot-taxi services seem to be possible and Advanced Driver Assistance Systems (ADAS) are becoming standard in new car models and even mandatory in the EU. But enabling a car to safely drive and navigate through complex environments is not an easy task. Every second on the road an autonomous vehicle must make several decisions based on the current road situation. Examples for such decisions are which lane to take or to how to approach other cars. The surrounding environment which is the basis for these decisions can be composed of a variety of elements, like static obstacles, other road participants, traffic signs and pedestrians. The ability to process such a quantity of information and use it to make appropriate decisions is currently one of the biggest challenges in the field of autonomous driving. Designing rules for all possible scenarios is an almost unfeasible task. Due to that all the current applications are either limited to rural areas with medium complexity or the function itself, like preceding a car is rather simple. The paper introduces a combination of machine learning and conventional control theory and how it can be used to safely navigate in different highway scenarios without explicit rules.

Using machine learning in the area of autonomous driving has become increasingly popular, most notably in camera perception. Almost every modern object recognition algorithm is powered through Deep Learning [1]. But machine learning can also be applied for end-to-end control systems, an example is a system which learns to produce a steering signal based on an image [2][3]. But a drawback of those techniques is that they require large amounts of labeled data to achieve good results. Creating such data for the complex decision-making processes in autonomous driving is almost impossible. Therefore, a system which could learn to deal with those decisions without explicitly labeled data or a predefined set of rules would be valuable. A machine learning technique with the potential solve this problem is Deep Reinforcement Learning (DRL).This combination of deep learning and reinforcement learning has already been applied to create systems that can play ATARI games [4], the board game Go [5][6] or chess [7] on superhuman level. But it's not only limited to the field of gaming. It has been applied in robotics and enabled robots to self-learn manipulation [8] and grasping tasks [9]. First attempts have also been made in trying to use DRL for lane changes [10] and intersection handling [11] as well as further specific applications [12].

Based on this research the paper introduces a deep reinforcement learning centered autonomous system for lane change maneuvers. The decision making and timing is dealt with by a DRL policy and the motion planning and control is executed using conventional methods. Through this the decision-making process is simplified and the comprehensiveness of control theory is combined with self-learning capabilities of reinforcement learning. This novel approach has been evaluated on several scenarios and with different state representations. The result is a generic applicable system which can safely drive in highway scenarios using a self-learned policy.

## II. Approach

For solving the task of driving on highway scenarios a system combining Deep Reinforcement Learning with control theory is applied. A policy which outputs target points is learned using Deep Q-Learning [4]. Lateral and longitudinal movement are dealt with using controller and trajectory planning. In the following the fundamentals of the used algorithm and the used method is introduced.

### A. Reinforcement Learning

Reinforcement learning (RL) is one subsection of machine learning and can be described as learning through trial and error. A standard reinforcement learning system consists of an agent and an environment. In this scenario the agent is the learner and tries to cope with the challenges of the environment. The agent tries to find a mapping between a state and an action which maximizes a reward signal. Reinforcement learning problems are usually modelled as Markov Decision Process (MDP) or it's extension Partially observable Markov Decision Process (POMDP). This framework is described by the tuple $(S, A, P, R)$ and the assumption of the Markov Property which states that the next state of an environment only depends on the immediate previous state and action. The variable S is the state space which contains every possible state $s$ of the environment. A state contains the full information about the environment at the current time. The action space $A$ describes every action $a$ which can be executed by the agent to impact the environment. In P the state transition probability $p(s_{t+1}|s_t, a_t)$ for every case is given. The reward function R gives a feedback for executed actions in form of a scalar number. In this framework the agent tries to learn a policy $\pi(a_t|s_t)$ which enables it to choose the optimal action with the goal of maximizing the expected discounted reward $G_t$ over time [13].

$$G_t = \sum_{k=0}^{\infty} \gamma^k R_{t+k+1} \quad (1)$$

In the real world the whole state of the environment is not always fully observable due to limitations of sensors. Often only an observation $o$, which contains a fraction of information about the full state, exists. For this case the POMDP replaces the state with the observation $o$ and extends it with an emission probability $p(o_i|s_i)$ which describes the transition from state to observation. A modification of classical Reinforcement learning is Deep Reinforcement learning in which the policy of an agent is represented through a neural network.

## B. Deep Q-Network

One popular algorithm to solve a reinforcement learning problem is Q-Learning [14][15] in which the agent tries to learn a function for the so-called Q-value. It is the expected future reward for taking an action in a given state and then following the current policy $\pi$. The respective policy is then to take the action which yields the highest reward in the future. In traditional Q-Learning this function $Q(s, a)$ is represented through a table which provides the Q-value for every state and action pair. This Q-function is learned by taking an action with the current policy and observe the next state $s_{t+1}$ and received reward $r_t$. In an iterative process the Q-Function is then updated using the following equation:

$$Q(s_t, a_t) \leftarrow (1 - \alpha)Q(s_t, a_t) + \alpha[r_t + \gamma Q(s_{t+1}, a_{\max})] \quad (2)$$

The intention is to use the observed reward $r_t$ to obtain a better approximation of the Q-value as it is an estimate of future reward. In (2) $\gamma$ is a discount factor which accounts for the fact that future rewards are more uncertain and $\alpha$ is the learning rate. In Deep Q-Networks (DQN) the Q-table is replaced through a deep neural network $Q_\theta(s, a)$ for which the parameters $\theta$ are trained using back-propagation [16]. As opposed to supervised learning, there is no ground truth given in a reinforcement learning problem. In a forward pass the current estimate $Q_\theta(s_t, a_t)$ of the network for the Q-value is calculated. By executing the respective action, the tuple $(s_t, a_t, r_t, s_{t+1})$ is obtained and stored in an experience buffer. But instead of updating single Q-values, the network parameters $\theta$ are tuned using a cost function.

$$J_\theta = \frac{1}{2}(Q_\theta(s_t, a_t) - (r_t + \gamma \max_{a_i} Q_\theta(s_{t+1}, a_{max})))^2 \quad (3)$$

In (3) the definition $r_t + \gamma(Q_\theta(s_{t+1}, a_{max}))$ is used as an approximation of a ground truth $y$ and the network is trained with the resulting cost $J_\theta$. But the fact that the same function is used to calculate estimation and ground truth causes the learning to be unstable. Every training step the network parameters change and due to that cannot be optimized to a fixed target. In order to stabilize the learning process several modifications of the original algorithm have been made. The first one is Double Q-learning [17] which uses a separate target network $Q_T$ to compute the ground truth. In this algorithm the weights of this network are kept static and are only synced with the parameters of the main network $Q_M$ periodically. Hence, the correlation between ground truth and estimation is decoupled. A further modification is Dueling DQN [18] for which the Q-value computation is split into an advantage of an action $A$ and a state value $V$. The separate computation has been proven to further stabilize the result.

## C. State Representation

The policy should be trained to drive safely and efficiently on highway scenarios. For this an environment has been developed which provides a variety of highway scenarios and can be modified with a wide range of parameters. The state space consists of position and velocity of all vehicles relative to the ego-vehicle. In the current configuration only the ego-vehicle is capable of lateral movement and therefore only the longitudinal velocity is included.

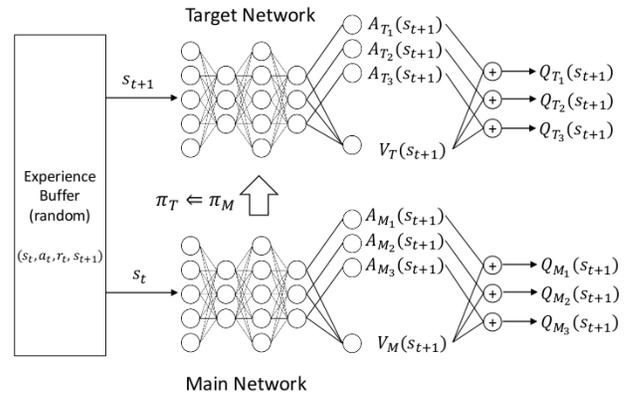

Figure 1: Combined structure of a dueling and double DQN

$$\vec{s}_i = \begin{pmatrix} x_i - x_{ego} \\ y_i - y_{ego} \\ v_i - v_{ego} \end{pmatrix} \quad (4)$$

The described state space does not include the position and velocity of the ego-vehicle which should ensure that the agent does not learn to react based on a driven distance but rather on the movement patterns around him. The state vectors of each vehicle are organized in a list with fixed size (case: fully observable).

In order to evaluate the performance of the system for partially observable cases and different network architectures two observations of this state space are taken. In one variant the maximum view ($d_x$) is limited because sensors have a maximum range (case: limited view). The second variant is a two-layer grid-map (size: $d_x \times d_y$) with the ego-vehicle as center. Processing the environment in a grid is common practice in autonomous driving. The position of the vehicles are the first layer and the velocity in the second (case: grid). Here the factory $d_y$ is determined through the This grid-observation is processed through a convolutional neural network [19]. For the described environment an own simulator has been implemented in order to keep the complexity low and make the training as efficient as possible.

### D. Action Representation

In theory Deep Q-Learning can be applied to a broad range of problems, however often well-formulated techniques from control theory can solve the problems more efficiently and without the black box characteristics of neural networks. Therefore, the previously described Deep Q-Learning algorithm is applied to output parameters for a trajectory planner. This should also enable a smoother transition from simulation to reality in the future as the prediction of a target is not affected by the internal physics of the environment. For this application the action space is defined as the following:

$$A = \{0, 1, 2, \ldots, L_{max} * D_{max} - 1, L_{max} * D_{max}\} \quad (5)$$

$D_{max}$ is a manually defined scalar factor and $L_{max}$ is the maximum numbers of lanes in the scenario. The respective

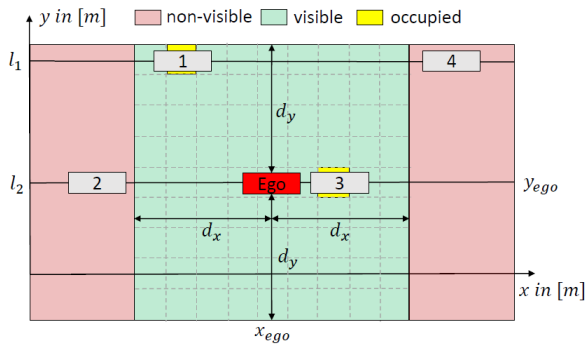

Figure 2: Grid representation of the environment

action $a_t$ is equivalent to a factor $d$ and a lane $l_i$. Those parameters are obtained by decoding the value of $a_t$ according to the following equations:

$$l_i = \text{round}\left(\frac{a_t}{D_{max}}\right) * w_{road} + \frac{1}{2} w_{road} \quad (6)$$

$$d = a_t \% D_{max} + 1 \quad (7)$$

This allows to use an integer number as coordinates on the map. Those values are the target coordinates for the trajectory planner. Based on this the start point $\vec{x_0} = (x_{ego}, y_{ego})$ and the target point $\vec{x_T} = (v_{ego} * d, l_i)$ are defined. Between those two points a quantic function with the parameters $\vec{p}$ is fitted using the definition of curvature as an optimization goal.

$$\kappa(x) = \frac{f''(x)}{(1 + f'(x)^2)^{\frac{3}{2}}} \approx f''(x) \quad (8)$$

Minimizing the curvature is equivalent to minimizing the lateral acceleration and therefore the optimization goal is:

$$\min_{\vec{p}} \int_{x_0}^{x_T} f''(x) \, dx = \min_{\vec{p}} \int_{x_0}^{x_T} \kappa(x) \, dx \quad (9)$$

Additionally, the constraints that the velocity and acceleration must be zero at the start and end point are defined. The parameters are then optimized using quadratic programming with constraints [20].

The agent follows the calculated trajectory but can decide to calculate a new trajectory at every timestep with a frequency of 5 Hz. It follows the trajectory without errors in order to provide a smooth learning process. This is a reasonable assumption as lateral movement can be sufficiently dealt with through control methods like Model Predictive Control [21]. The intention behind multiple longitudinal distances is to provide a possibility to adapt the lane change behavior according to the situation. In fig. 3 trajectories for different factors and their respective lateral acceleration is shown. The longitudinal motion is dealt with through a simple control algorithm which tries to keep a certain safety distance $d_s$ and the velocity of the car in front $v_{front}$. It follows the equation:

$$\text{acc}_x = \alpha(v_{front} - v_i) + 0.25\alpha^2(dist - d_s) \quad (10)$$

The controller is inspired by the widely used Intelligent Driver Model [22] but provides a simpler equation which still ensures safe longitudinal movement. In this setting the Deep

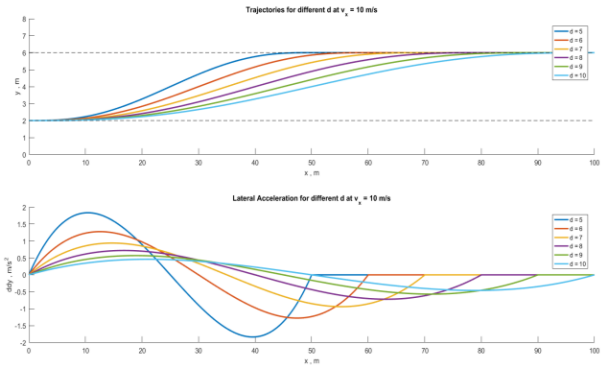

Figure 3: Sample trajectories for a lane change depending on the distance factor d

Q-Network is then trained to make appropriate lane change decisions.

## III. RESULTS

The designed approach of using a Deep-Q-Network as the central decision-making unit and supported by conventional control theory as described in section II is evaluated on two different scenarios. One is a simple overtake maneuver and another one includes on-coming traffic. For each scenario the training process as well as quantitative and qualitative results are analyzed. The system has been tried with three different inputs, which are the state (fully observable), a limited view and a grid map. In the overtake scenario the agent should take over multiple vehicles which drive slower in front of it and go back on the initial lane. In the second scenario on-coming traffic is added and the agent should ensure he is not blocking or crashes into other vehicles. The trained policies are evaluated over multiple random seeds to provide a statistical more relevant result. In deep reinforcement learning it's a common issue that results strongly vary depending on the random seed and implementation [23].

### A. Set-Up

In this section the training process for the two cases are analyzed. As described previously the system is evaluated using three different inputs and for two scenarios. In table I the parameters for these scenarios are provided. The non-ego vehicles are randomly initialized in front of the ego vehicle and execute their lateral movement as described in section II.D The network architecture for the policy uses the same number of units for the hidden layers and the input- and output layer is modeled according to the state and action space. The full set of reinforcement learning, and neural network parameters are given in table II. The parameters were defined using a grid search during preliminary tests and thereby analyzing the impact of the different parameters. For the reward, a function specially defined for highway scenarios has been used. It considers the velocity of ego- and non-ego vehicles, lateral acceleration and completion of the episode.

TABLE I. SCENARIO PARAMETERS FOR TRAINING

| Scenario parameters | |
|---|---|
| Non-Ego Cars | 2 |
| Lanes | 120 km/h (33.33m/s) |
| Initial ego velocoty | 120 km/h (33.33m/s) |
| Non-ego velocity limit | 84 km/h |
| Range of view | 150 m |
| Initation range | 300 m |
| Maximum Duration | 70 s (450 steps) |

For the reward a function specially defined for highway scenarios has been used. It considers the velocity of ego- and non-ego vehicles, lateral acceleration and completion of the episode. The reward $r(s)$ has been defined as:

$$r_1 = \ddot{y}^2 * 0.12 - (v_{max} - v_{ego}) * 2 - \ddot{x}^2 \quad (11)$$

This part encourages behavior which maximizes the velocity of the ego-vehicle and avoids unnecessary accelerations.

$$r_2 = 1.375 * dist_{lat}^2 - 6.25 * dist_{lat} + 5 \quad (12)$$

The equation $r_2$ has been designed to allow lateral movement but to encourage to reach the lane and not to stay in between.

$$r_3 = -0.1 * mean(v_{non}^2) \quad (13)$$

In this component a high velocity of every non-ego vehicle is rewarded. This should encourage behavior of the agent to not slow down other vehicles. The whole reward function is composed out of those single equations and additionally driving on the centers of the lane is rewarded. If the scenario is completed successfully the agent receives an additional increment of 1000.

### B. Training

In the following the training process for the different agent systems trying to learn a policy for each scenario is displayed and analyzed. In fig. 4 and fig 5 the development of the reward, average velocity and number of completed episodes for both scenarios during training is displayed. Every 200 steps the performance is evaluated on 100 random episodes. For all cases the confidence interval across the random seeds are shown. For both scenarios the agents should learn to complete the episodes and thereby increase their average velocity. It is observable that for both scenarios the list-based approaches show the desired behavior although with strong oscillations, especially in the second scenario. This oscillation can be explained with the fact that the Deep Q-Network can't perfectly converge because no static optimization target is provided as explained in section II.B.

TABLE II. NETWORK AND RL PARAMETERS

| Training parameters | |
|---|---|
| Episodes | 2 |
| Random seeds | 3 |
| Layers | 3 |
| Hidden units | 50 |
| Learning rate | 0.001(Adam) |
| Discount factor | 0.99 |
| Exploration factor $\varepsilon$ | 1-0.1 |
| $\varepsilon$ steps | 150000 |
| Update frequency | 20000 |
| Batch size | 32 |
| Buffer size | 100000 |

However, for the first scenario, given the static increase across all parameters, it can be stated the list-based approaches can learn to solve the given task in an efficient manner. In the second scenario this behavior is less obvious, due to the strong oscillation. But looking at peaks during later episodes it is also visible that good performance metrics are achieved. In comparison the grid-based approach does not reach the same rewards or number of completed episodes, especially for the first scenario. But on the other side achieves a, more stable, less oscillating result. When analyzing the behavior, it becomes clear that the learned policy for this agent shows a more passive behavior and doesn't dare to complete the overtake. Given the design of the scenarios through this it is not possible to complete the scenario, but it can still be considered a safe behavior.

Regarding the training two observations can be made. First an agent with limited view is outperforming its counterpart with a complete view of the environment. It can be concluded that only considering the nearest non-ego vehicles into the decision-making process can improve the results. Furthermore, the grid-based agent, even though not reaching the same performance as the two others, delivers the most stable results with the smallest variance in the results. Overall the analysis shows a Deep Q-Network can learn the task of computing target positions in the given framework and thereby control the decision making in the car.

## C. Performance

In the following the quantitative and qualitative results of the trained policy are analyzed. Given the oscillating training process in deep reinforcement learning it is common to pick the best performing policy rather than the final one after completing the training. This strategy is similar to early stopping for neural nets trained in supervised learning [24]. The selection process is based on the achieved reward, average velocity and completed episodes. Each agent is tested for 100 episodes. Results for the first scenario are given in table III. In brackets the average across all seeds is provided.

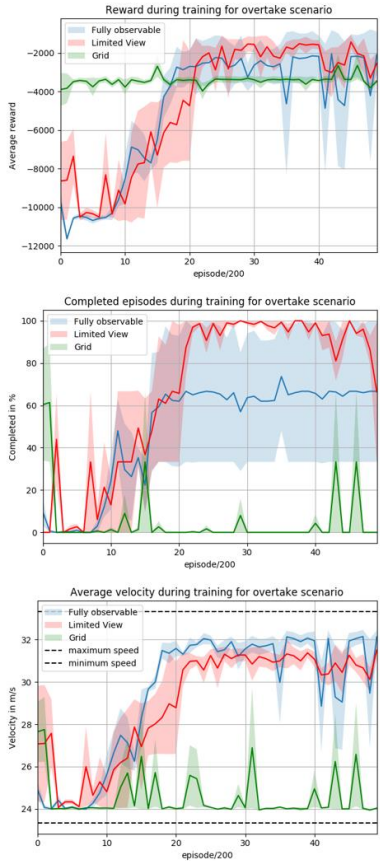

Figure 4: Training process for the overtake scenario

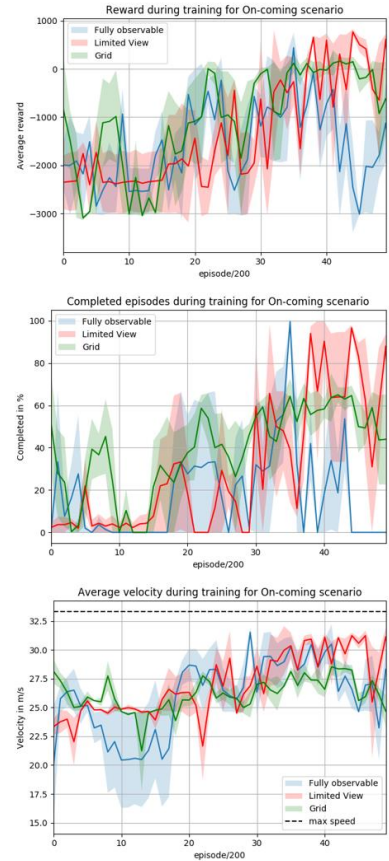

Figure 5: Training process for the oncoming traffic scenario

Across all used observations the best performing policy can solve one hundred percent of the overtake scenarios. As already seen in the analysis of the training process the agent with limited view outperforms the agent using the fully observable state. Also, it's average velocity almost reaches the maximum speed because the system can react to the slow down caused by the vehicles in front. The reason why the maximum speed is not reached is due to ending the episode early. Even though the training of the grid agent is not on the same level as the two list-based approaches it still is able to solve well but with bigger divergence across random seeds.

In the on-coming scenario the possibility of collision through wrong lane change decision is given and therefore is also considered when analyzing the systems performance. In table IV the respective numbers are provided. It shows that the combined system of machine learning and control engineering can completely solve the given task. In the best case the agent drives through the scenario without any collisions and with almost maximum speed. The agent with limited view as in the previous scenario has the best completion rate and velocity. Only regarding collisions, the agent with fully observable state has a slight edge. But this is expected because it can't be surprised through hidden non-ego vehicles. The grid-based agent is not able to solve all given episodes but also doesn't produces any collision.

This further supports the assumption that the use of grids and convolutional neural networks results in more conservative behavior. Finally, when observing the generated trajectories of the agents in case of a safe traffic situation no unnecessary interruptions of the lane change are generated (see fig 6.). The agent can make multiple decisions during a lane change and the shown behavior indicates that it has the right timing as well as an understanding of how to efficiently execute. One example of a successful overtake is shown in fig. 7. It shows the agent (red car) initiates the lane change in time, overtakes the car in front while staying on the adjacent lane and returns to the initial lane after passing the respective cars. For the scenario with on-coming traffic the agent also interrupts its lane change when the risk of a collision is given due to on-coming traffic.

TABLE III. QUANTITATIVE RESULTS FOR THE FIRST SCENARIO

| Quantitative results for overtake scenario | | | |
|---|---|---|---|
| *Observation* | *Completion %* | *Avg. reward* | *Avg. velo in m/s* |
| Fully observable | 100 (73.66) | -857 (-4639) | 31.97 (29.97) |
| Limited view | 100 (100) | -1022 (.1432) | 31.88 (31.26) |
| Grid | 100 (61.33) | -1766 (-3835) | 30.48 (27.75) |

TABLE IV. QUANTITATIVE RESULTS FOR THE FIRST SCENARIO

| Quantitative result for on-coming scenario | | | |
|---|---|---|---|
| *Observation* | *Completion %* | *Collision %s* | *Avg. velo in m/s* |
| Fully observable | 100 (66.6) | 0 (0) | 30.78 (28.5) |
| Limited View | 100 (100) | 0 (1.6) | 31.84 (31.23) |
| Grid | 75 (65) | 0 (9.33) | 30.03 (31.24) |

### D. Discussion

It must be mentioned that for all cases the reward function strongly affects the results. A careful design of this function is a key in achieving satisfying performance. For example, if the lateral acceleration is not penalized accordingly, the ego-vehicle drives between two lanes. A research direction that deals with this problem is inverse reinforcement learning in which the reward function is learned based on demonstration. There have been promising results using human driving data to learn reward functions for different scenarios [25] [26]. Another important remark is that the policy learned with this method is only able to cope with situations it's exposed to during training. Therefore, a careful design of the simulation in order to cover all possible situations for the desired application must be ensured. Currently the system is also limited to non-lateral movement scenarios and needs to be trained in more complex scenarios. In order to successfully deal with this the used state needs to be extended by orientation parameters in order to discover lane change intentions of non-ego vehicles.

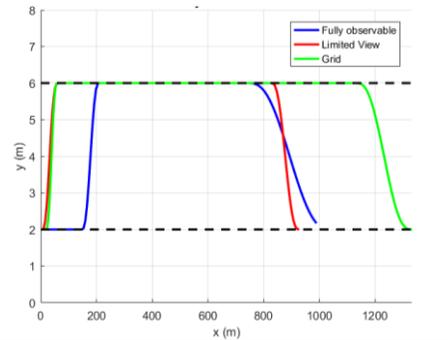

Figure 6: Sample trajectories generated by the different agents

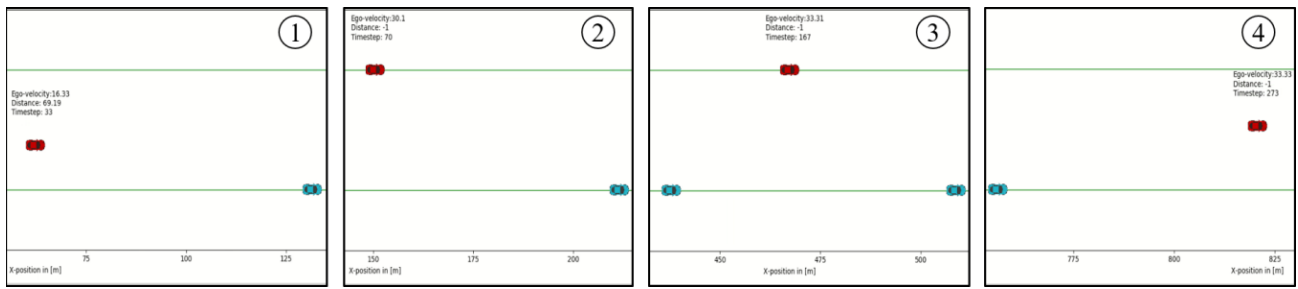

Figure 7: Example of a successful overtake maneuver (red: ego-vehicle)

Nonetheless, the analysis of the results shows that the proposed approach of using a Deep Q-Network for decision making can solve highway scenarios efficiently and without collisions. It was able to learn appropriate lane change behavior without explicit definition of rules. Using Deep Q-Networks to predict target points for trajectory planning can serve as a new method to deal with the challenges in decision making and planning for autonomous driving. It allows to add safety measures and simplify the process by restricting the target points to the possible lanes. It can be assumed that the approach provides the possibility to achieve a smooth transition from simulation to real world as the prediction of target points is not significantly influenced by internal physics of the environment. The control part of the system is used in order to deal with those constraints. Another possible application could be the combination with online-trajectory search methods. In this case the policy could be used as an underlying heuristic to accelerate the search process. Finally, this method is generally applicable and could also be applied for intersections and more complex traffic scenarios.

IV. CONCLUSION

In this paper a general applicable approach has been introduced which uses a Deep Q-Network as central decision and planning unit in autonomous driving. It has been shown that this system can be trained to calculate target points in order to drive efficiently and collision-free on highway scenarios as well as that considering only the vehicles in close range to the car can improve the results. Topics for future research include the use of more complex state- and action spaces to deal with the limitations, design of the reward, combination with online planning and the step from simulation to environment which could further improve the results as well as the capabilities of autonomous vehicles.

ACKNOWLEDGMENT

We would like to thank the team of BMW's connected and automated driving lab for the support during the process of this research project.